\documentclass[10pt,twocolumn,letterpaper]{article}

\usepackage{cvpr}
\usepackage{times}
\usepackage{epsfig}
\usepackage{graphicx}
\usepackage{amsmath}
\usepackage{amssymb}

\usepackage{algorithm}
\usepackage{algorithmic}
\usepackage[noblocks]{authblk}
\usepackage{subfigure}
\usepackage{multirow}

\usepackage[breaklinks=true,bookmarks=false]{hyperref}

\cvprfinalcopy 

\def\cvprPaperID{1868} 

\ifcvprfinal\fi
\begin{document}

\title{Top-push Video-based Person Re-identification\thanks{Citation for this paper: Jinjie You, Ancong Wu, Xiang Li and Wei-shi Zheng. “Top-push Video-based Person Re-identification.” In IEEE CVPR, 2016.}
}

%

\author[$\S$,$\ddag$]{\vspace{-0.5cm}Jinjie You}
\author[$\S$,$\ddag$]{Ancong Wu}
\author[$\S$,$\ddag$]{Xiang Li}
\author[$\S$,$\dag$,$\ddag$]{Wei-Shi Zheng\thanks{Corresponding author}\vspace{-0.3cm}}

\affil[$\S$]{Intelligence Science and System Lab, Sun Yat-sen University, China}
\affil[$\dag$]{School of Data and Computer Science, Sun Yat-sen University, China\vspace{0.0cm}}
\affil[$\ddag$]{Guangdong Provincial Key Laboratory of Computational Science, China}

\affil[ ]{\tt\small youjinjie9@gmail.com, wuancong@mail2.sysu.edu.cn

lixiang651@gmail.com, wszheng@ieee.org}

\maketitle
\thispagestyle{empty}
\newcommand{\modify}[1]{{\color{black}#1}}

\begin{abstract}

Most existing person re-identification (re-id) models focus on matching still person images across disjoint camera views. Since only limited information can be exploited from still images, it is hard (if not impossible) to overcome the occlusion, pose and camera-view change, and lighting variation problems. In comparison, video-based re-id methods can utilize extra space-time information, which contains much more rich cues for matching to overcome the mentioned problems. However, we find that when using video-based representation, some inter-class difference can be much more obscure than the one when using still-image-based representation, because different people could not only have similar appearance but also have similar motions and actions which are hard to align. To solve this problem, we propose a top-push distance learning model (TDL)\footnote{Our code is available at \href{URL}{http://isee.sysu.edu.cn/files/resource/TDL.zip.}}, in which we integrate a top-push constrain for matching video features of persons. The top-push constraint enforces the optimization on top-rank matching in re-id, so as to make the matching model more effective towards selecting more discriminative features to distinguish different persons. Our experiments show that the proposed video-based re-id framework outperforms the state-of-the-art video-based re-id methods.

\end{abstract}

\section{Introduction}

Person re-identification (re-id) matches persons across non-overlapping camera views at different time. Most existing works focus on matching still images represented by appearance features (\eg color histograms), because of computation efficiency and limited storage space. 
Given a probe image, we match it against a set of gallery images, which may suffer from illumination change, viewpoint difference, complicated background and occlusions. The significant visual ambiguity and appearance variation make still-image-based person re-id a challenging problem. Many methods have been developed to  either extract invariant features or learn discriminative matching models \cite{Gheissari06a, WangShapeAppearance_07a, gray2008viewpoint, farenzena2010person, ma2012local, li2015partial, kviatkovsky2013color, prosser2010person, Zheng_2011_PRDC,mignon2012pcca, koestinger2012large,zhao2013unsupervised,li2013learning, pedagadi2013local,li2013locally,Zhao_MidLevel_2014a,lisanti2014person, ma2014person, xiong2014person,li2014deepreid,ahmed2015improved,liao2015person,wang2015cross,chenasymmetric}.

\begin{figure}[t]
\begin{center}
  \includegraphics[width=0.9\linewidth,height=0.5\linewidth]{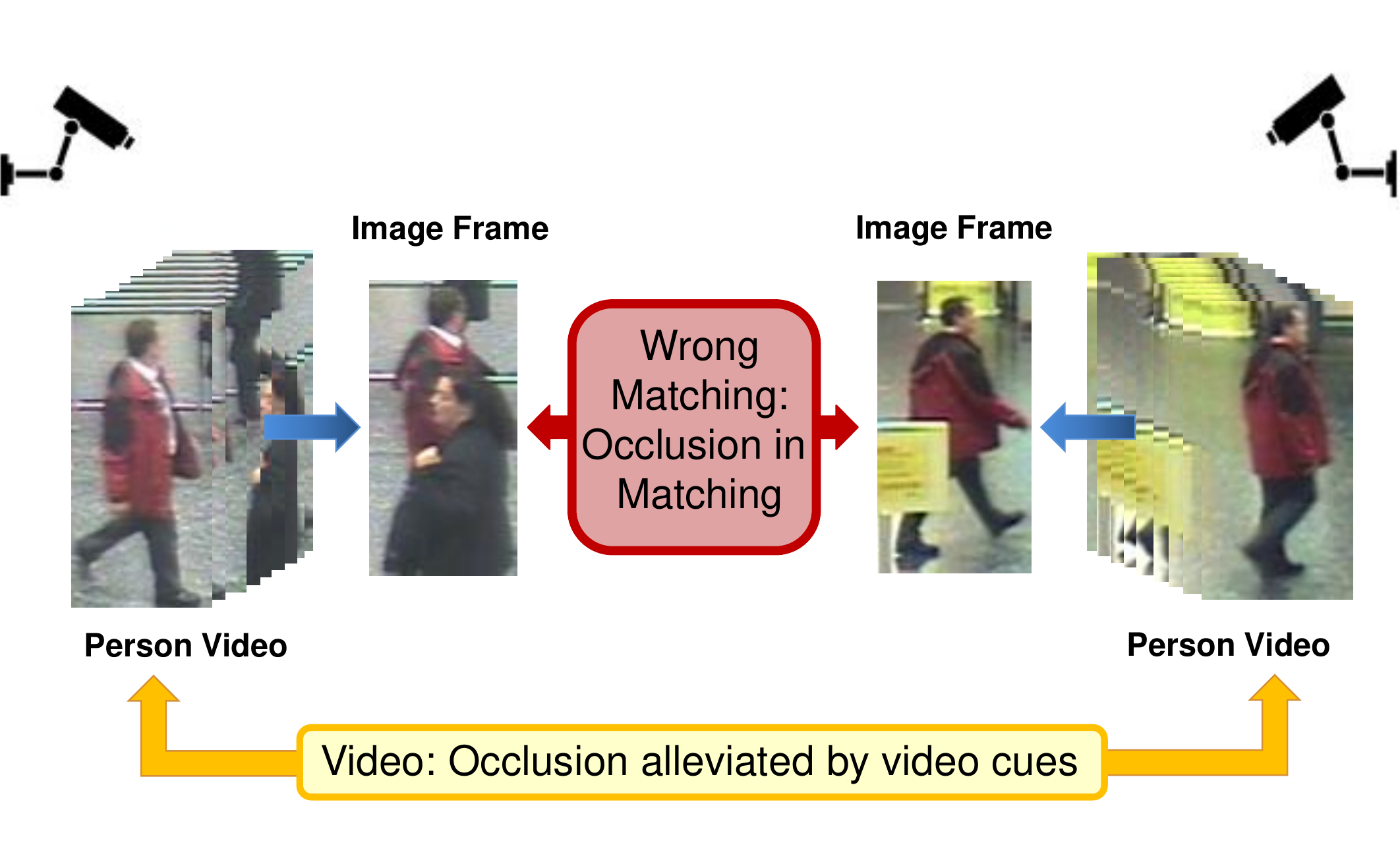}
\end{center}\vspace{-0.5cm}
   \caption{Video instances vs. still-image instances of the same person. It is clear that video contains much richer cues for matching.}
\label{fig:IMvsVID}\vspace{-0.3cm}
\end{figure}

\begin{figure*}[t]
\centering{
\subfigure[ ]
{
     \includegraphics[width=0.9\linewidth,height=0.17\linewidth]{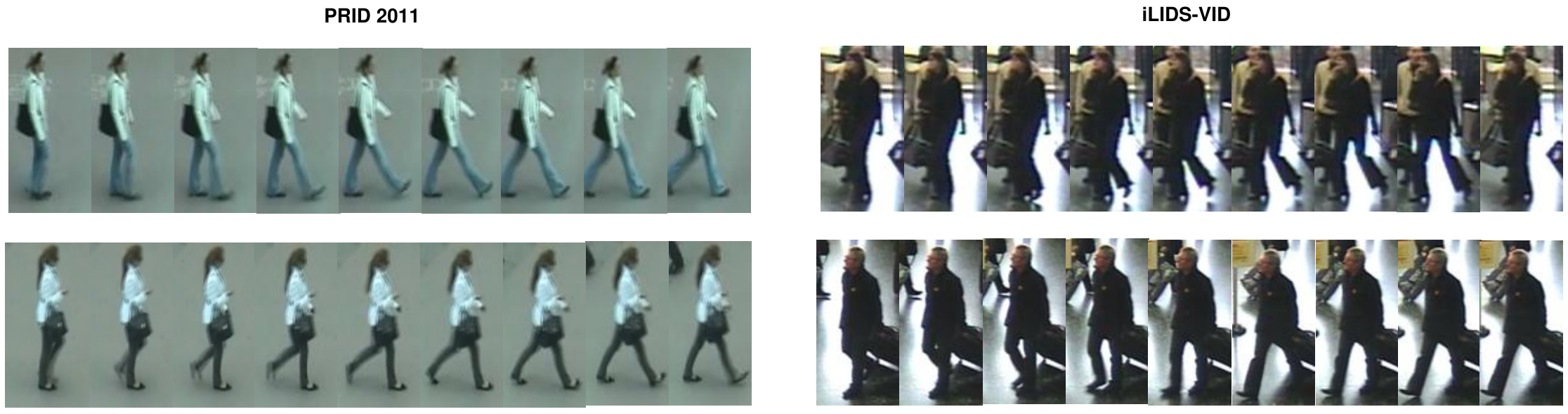}
}
\subfigure[ ]
{
    \includegraphics[width=0.9\linewidth,height=0.14\linewidth]{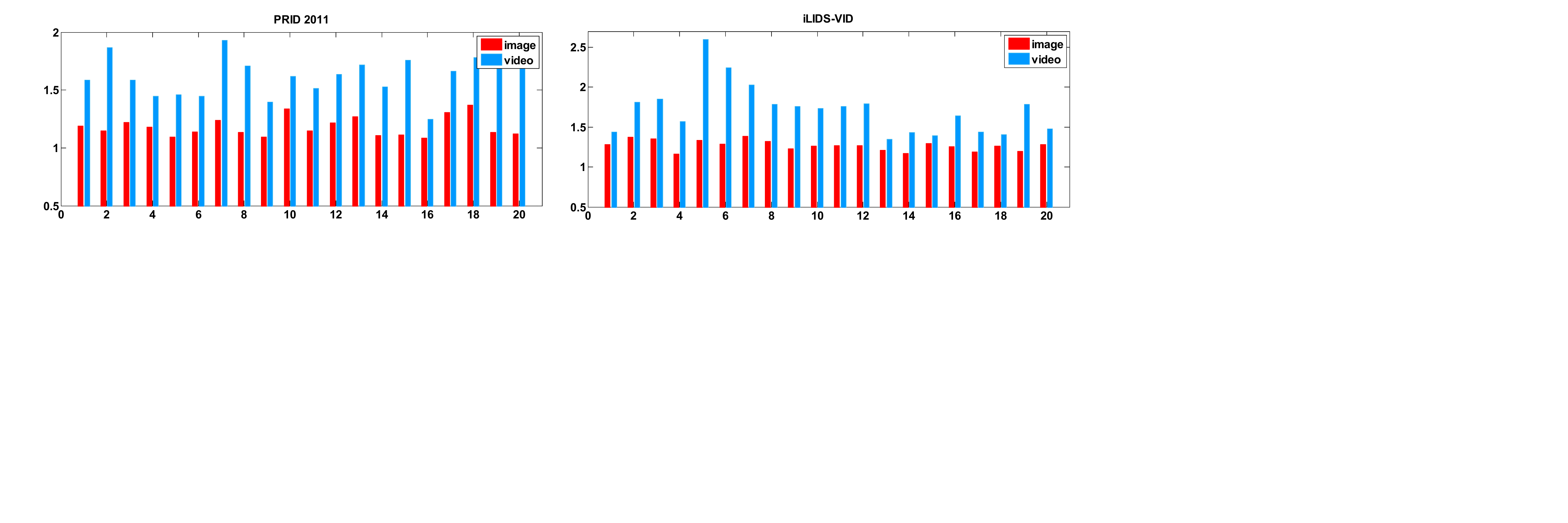}
}
}  
\caption{In (a), on each dataset, we show two video instances of different people, who are wearing similar clothes and walking similarly. In (b), we randomly selected 20 video instances of different people. For each instance, we compute its image frame level feature representation (color\&LBP) and video level's (HOG3D+Color\&LBP(pooling)) where for the image level's we randomly selected one image frame from a video instance. For each level's representation, we compute the largest intra-class distance $D_w$ and the smallest inter-class distance $D_b$ with respect to each sample.
The x-axis is the index of the 20 random samples and the y-axis is the value of $D_w/D_b$. It can be observed that most ratio values of videos are larger than those of image frames, \ie these videos have more ambiguities than images.}
\label{fig: distance_bar}\vspace{-0.3cm}
\end{figure*}

However, the still-image-based person re-id indicates that the temporal information between images of a person in each camera view is ignored. Information of a still image is sometimes not enough for recognizing a person, \eg the person being occluded by objects or other persons (see Figure \ref{fig:IMvsVID} for example). As surveillance information is recorded by videos and human operators always recognize persons in videos, it is intuitive to mine more effective information in video re-id. What more information can we obtain from videos than still images? Firstly, video is an image sequence containing space-time information, in which motion information is available. Secondly, appearance cues are more abundant in a sequence than in a still image, which can facilitate extracting more robust appearance features. Thirdly, occlusion and background influence can be eliminated to some extent. In a sequence, background variation and occlusion can be regarded as removable noises, while in still images they are troubling interferences.

Although more information can be obtained from person videos, more challenges come along. Firstly, like the still-image-based approaches, video-based person representations are also similar because of similar appearance. Secondly, although the motion of a pedestrian is a kind of behavioral biometrics, that is an important discriminative cue for identifying different persons, it is unfortunate that the walking actions or other motions of different persons may be similar as well (see Figure \ref{fig: distance_bar} (a) for example), which means the inter-class variation may be smaller for video-based representation of a person. As shown in Figure \ref{fig: distance_bar} (b), we demonstrate that for some instances, it is harder to distinguish the video representations of different identities (due to large $\frac{D_w}{D_b}$ value) than the still image cases.
\modify{It shows that the ratio between maximum intra-class distance and minimum inter-class distance is much larger for the video-based re-id as compared to the image-based re-id because some motion information of different people could be similar. This suggests the ambiguity of videos is more serious, and it is true that more intra-class distances are larger than the related minimum inter-class distance.}
The observation here would imply the discriminative information could be hidden in the minor difference of actions and motion. To mine these minor differences in the data, more stringent constraint should be exploited to look for a latent space to maximize the inter-class margin between different persons. So far, only a few video-based methods \cite{wang2014person, karanam2015sparse,karanam2015iccv} have been developed. However, the mentioned problem for video-based person re-id still remains unsolved.

To address the above problem in video-based person re-id, we propose a top-push distance learning model (TDL) in this work. For a person video sequence, we exploit a feature representation constituted by HOG3D \cite{klaser2008spatio} and the average pooling of color histograms and LBP features \cite{hirzer2012relaxed}. Based on that, we propose a discriminative distance model optimized towards the realization of the top-push distance constraint combined with the minimization of intra-class variations.
We employ the idea of top-push in \cite{li2014top} and introduce it into distance metric learning, in order to optimize the matching accuracy at the top rank for person re-id, which helps to look for a latent feature space to explicitly enlarge the inter-class margin between video sequences.






Extensive experiments have been conducted on two video datasets including PRID 2011 \cite{hirzer2011person} and iLIDS-VID \cite{wang2014person} to validate the effectiveness of the proposed TDL model. Our results demonstrate that
(1) by formulating the video-based person re-id problem as a
distance metric learning problem with top-push constraint modeling, significant improvement on matching accuracy
can be obtained against the existing video-based person re-id techniques; and (2) our proposed TDL model outperforms not only related distance/rank learning methods but also related representative still-image-based person re-id methods applied for the video-based person re-id problem under multi-shot setting.

\section{Related Works}\label{section:related_work}

The unsolved problem of person re-id caused by lighting change, viewpoint change, occlusions and intricate background has been increasingly focused on and becomes an important topic in visual surveillance in the last five years.
To overcome these challenges, most of existing works can be categorized into extracting discriminant/relible features \cite{Gheissari06a, WangShapeAppearance_07a, gray2008viewpoint, farenzena2010person, ma2012local, kviatkovsky2013color, wuviewpoint,shangxuan2016wacv} or learning robust metrics or subspaces for matching \cite{yingcong2015mirror, prosser2010person, Zheng_2011_PRDC, mignon2012pcca, koestinger2012large, li2013learning, pedagadi2013local, ma2014person, xiong2014person, liao2015person, liao2015psdre-id} in recent works.
However, all these works use appearance features of still images to match, which may suffer from small inter-class variations caused by similar pedestrians clothing and large intra-class variations caused by occlusions. Although it is natural to extend them to handle video-based person re-id under a multi-shot setting, it is not an optimal way as shown in our experiments, moreover it takes more times for matching due to the increase of gallery size.

Recently, a few works started to consider solving the person video matching problem in re-id.
Dynamic Time Warping (DTW), which is a popular sequence matching method widely used for action recognition \cite{lin2009recognizing}, was applied for video-based person re-id \cite{simonnet2012re}.
Wang \etal  \cite{wang2014person} introduced a pictorial video segmentation approach and deployed a fragment selecting and ranking model for person matching.
Srikrishna \etal  \cite{karanam2015sparse} introduced a block sparse model to handle the video-based person re-id problem by the recovery problem on embedding space. However, these works assume all image sequences are synchronized, but it becomes unapplicable due to different actions taken by different people. It is also costly and difficult to obtain perfectly aligned pairwise person videos across non-overlapping camera views.
All these works use either multiple images or a selected fragment of a sequence to extract feature, and thus they ignore the integrity and the richness of video features. So they are ineffective for solving the video-based person re-id problem.

\begin{figure}
\centering{
\subfigure[LMNN]
{
   \includegraphics[width=0.42\linewidth,height=0.4\linewidth]{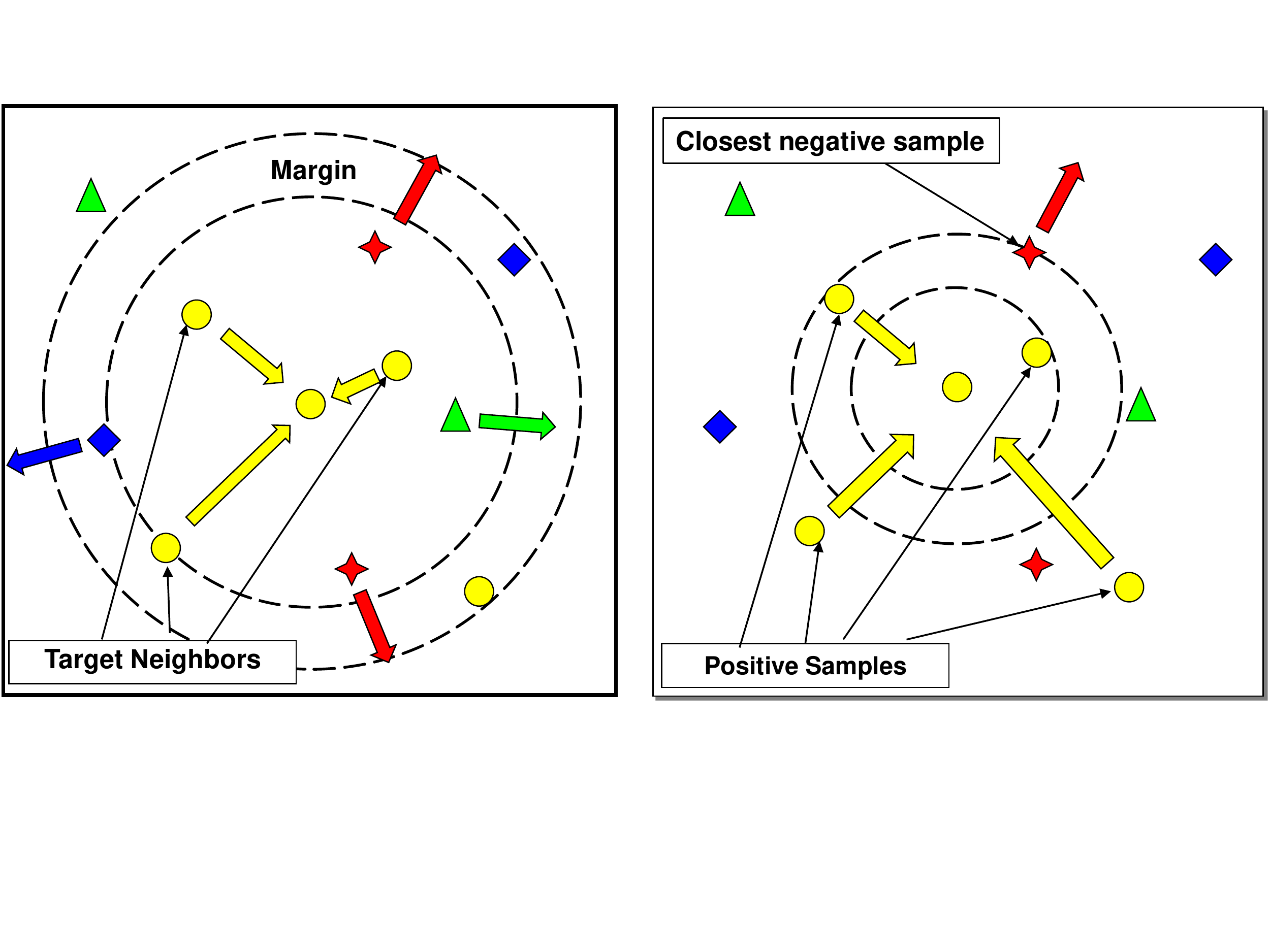}
}
\subfigure[TDL]
{
   \includegraphics[width=0.42\linewidth,height=0.4\linewidth]{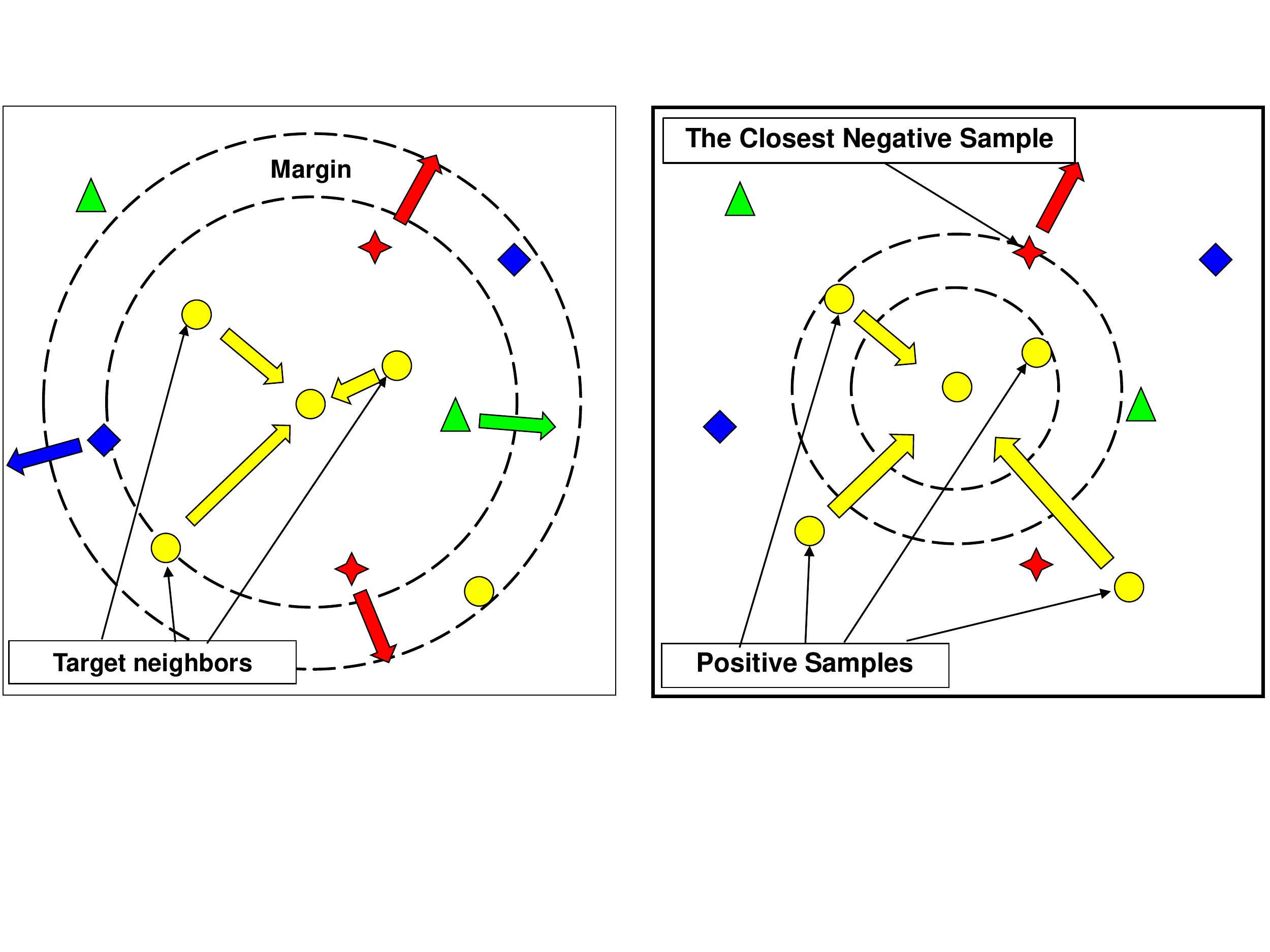}
}}
   \caption{LMNN (left) vs. our TDL (right). Compared to LMNN, since the minimum inter-class distance is considered in TDL, the imposters are more heavily penalized.}
\label{fig:toppush}\vspace{-0.2cm}
\end{figure}

We extend the use of top-push constraint from linear ranking function \cite{li2014top} to second-order distance metric learning in our TDL model. Both the proposed TDL model and the linear function in \cite{li2014top} aim to optimize the top-rank matching performance. The difference is that the TDL model is able to look for a latent subspace rather than computing only one ranking function score, so that more robust latent features can be exploited. Since, the top rank linear function learning is a RankSVM \cite{prosser2010person} like learning, which has been shown very costly on high dimensional and moderately large-scale dataset \cite{zheng2015verification}, the top rank linear function learning cannot be straightforward generalized to a multiple dimensional one \cite{li2014top}. Our experiments suggest that exploring a subspace rather than a hyperplane is more robust for person re-id.

Different from existing distance metric learning methods, our proposed TDL model is specially motivated from the observation that the inter-class variation is much smaller on video level than that on still image level, so the top-push constraint, a more effective relative comparison, is employed to explicitly avert this problem in a latent feature space.
In particular, our approach is related to Weinberger \etal's LMNN method \cite{weinberger2009distance}. LMNN aims at optimizing KNN classification by using the local structure of the data. For each instance, a local neighborhood is established, including the $k$ nearest neighbors sharing the same label (target neighbors). Samples that invade this perimeter with a different label (impostors) are penalized (see Figure~\ref{fig:toppush}~(a)). Our method seems similar to LMNN; however, an important difference is that a more stringent top-push constraint is used to guide the distance learning, which notably benefits the top-rank matching results in person re-id (see Figure~\ref{fig:toppush}~(b)). Ours is also related to the relative distance comparison (RDC) \cite{zheng2013reidentification}. While RDC is limited by the scale of relative comparison, the proposed TDL can largely reduce the number of relative comparisons in the context of top-push modeling. In addition, compared to LDA \cite{fukunaga2013introduction}, our model replaces the maximum of inter-class distance by the minimization of hinge loss of top-push comparison, so that our model has imposed much more powerful constraint on the inter-class modeling. The significant improvement against LMNN, RDC and LDA will be shown in the experiment part.




\section{Approach}

The feature representation of a person video in our model has two main components: space-time features and appearance features.
For extracting the space-time features, we employ the HOG3D descriptor \cite{klaser2008spatio} to represent the person video. The HOG3D feature contains spatial gradient and temporal dynamic information.
For extracting the appearance features, we first use color histograms and LBP features \cite{hirzer2012relaxed} to describe a person appearance in each image frame. To obtain stable appearance cues and suppress noises caused by occlusions, we express the appearance features of a person video by average pooling of features of all frames from that video. The average pooling of color histograms and LBP features can represent rich appearance information of a person in video.
The space-time features and the appearance features describe different information of a person in video, and those two types of features are complementary. Therefore in our model, the two features are combined to address the challenging video-based person re-id problem caused by background change, occlusions and motions.

We denote the training set by $X=\{(\vec{x}_i,y_i )\}_{i=1}^{s}$, where $\vec{x}_i\in\mathbb{R}^d$ is the feature vector extracted from a video of person labeled $y_i$. We denote the distance between any two feature vectors $\vec{x}_i$ and $\vec{x}_j$ by $\mathcal{D}(\vec{x}_i,\vec{x}_j)$.

\subsection{Enhancing Top-rank Matching by Top-push Distance Learning}

For person re-id, it is always expected that, for a query image, the top-rank matching of gallery images is correct. This means the distance between any matched gallery sample and the query should be smaller than the one between any unmatched one and the query. Therefore, in our distance metric learning modeling, we are concerning the relative comparison between the distance of a positive pair and the minimum distance of all related negative pairs, rather than comparing the positive pair with each of the related negative pair. In formulation, that is, for each example $\vec{x}_i$, we wish to realize the following comparison:
\begin{equation}\label{eq:top_push_comparison}
\mathcal{D}(\vec{x}_i,\vec{x}_j) + \rho < \min_{y_k \neq y_i} \mathcal{D}(\vec{x}_i,\vec{x}_k), \ y_i = y_j,
\end{equation}
where $\rho$ is a slack parameter. In this work, we set $\rho = 1$. To quantify the above comparison, we aim to minimize a hinge loss function incurred by the positive pairs whose distances are not smaller than the smallest distance of negative pairs with respect to input $\vec{x}_i$:
\begin{equation}\label{eq:hingeloss}
\min \sum_{\vec{x}_i,\vec{x}_j,y_i=y_j} \max\bigl\{\mathcal{D}(\vec{x}_i,\vec{x}_j) - \min_{y_k \neq y_i} \mathcal{D}(\vec{x}_i,\vec{x}_k) + \rho, 0 \bigr\}.
\end{equation}
The minimization of the loss of the above comparison refers to inter-class separation, which however does not address the intra-class variation. Therefore, we also wish to strengthen the correlation of samples of any positive pair by minimizing the distance between samples of the same class in the meanwhile, i.e.,
\begin{equation}\label{eq:hingeloss}
\min \sum_{\vec{x}_i,\vec{x}_j,y_i=y_j} \mathcal{D}(\vec{x}_i,\vec{x}_j).
\end{equation}
Therefore, the objective function of top-push distance learning is formulated below:
\begin{equation}\label{eq:toppush_criterion}
\begin{split}
f(D)=&(1-\alpha)\sum_{\vec{x}_i,\vec{x}_j,y_i=y_j} \mathcal{D}(\vec{x}_i,\vec{x}_j)\\
+\alpha & \sum_{\vec{x}_i,\vec{x}_j,y_i=y_j} \max\bigl\{\mathcal{D}(\vec{x}_i,\vec{x}_j) - \min_{y_k \neq y_i} \mathcal{D}(\vec{x}_i,\vec{x}_k) + \rho, 0 \bigr\},
\end{split}
\end{equation}
where $\alpha\in[0,1]$ refers to a weighting parameter that balances the two terms. We call the second term the \emph{top-push constraint}.
Through the optimization, the first term penalizes large distances between positive pairs, and meanwhile the second term penalizes small distances between each sample and the closest sample that is differently labeled. The learning induced by this cost function are illustrated in Figure~\ref{fig:toppush} for an input. We call our approach the top-push distance learning (TDL). In TDL, we specially consider the optimization of Mahalanobis distance under Criterion \ref{eq:toppush_criterion}, i.e., considering
\begin{equation}\begin{aligned}
\mathcal{D}(\vec{x}_i,\vec{x}_j)=(\vec{x}_i-\vec{x}_j)^{\top}\mathbf{M}(\vec{x}_i-\vec{x}_j),
\end{aligned}\end{equation}
where $\mathbf{M}\succeq0$ is a positive semi-definite matrix.

\begin{figure}[t]
\centering{
\subfigure[Before TDL]
{
   \includegraphics[width=0.47\linewidth,height=0.4\linewidth]{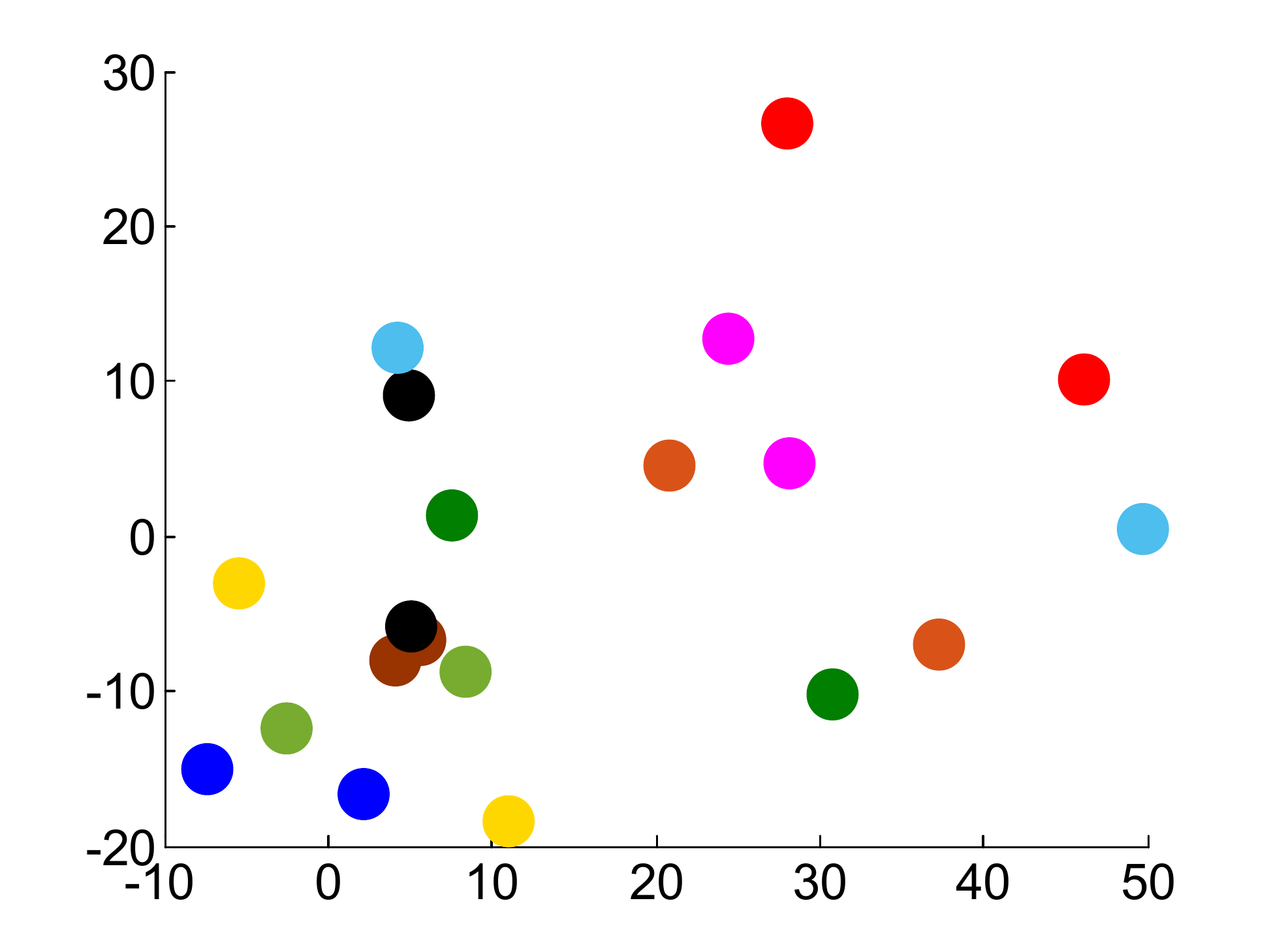}
}
\subfigure[After TDL]
{
   \includegraphics[width=0.47\linewidth,height=0.4\linewidth]{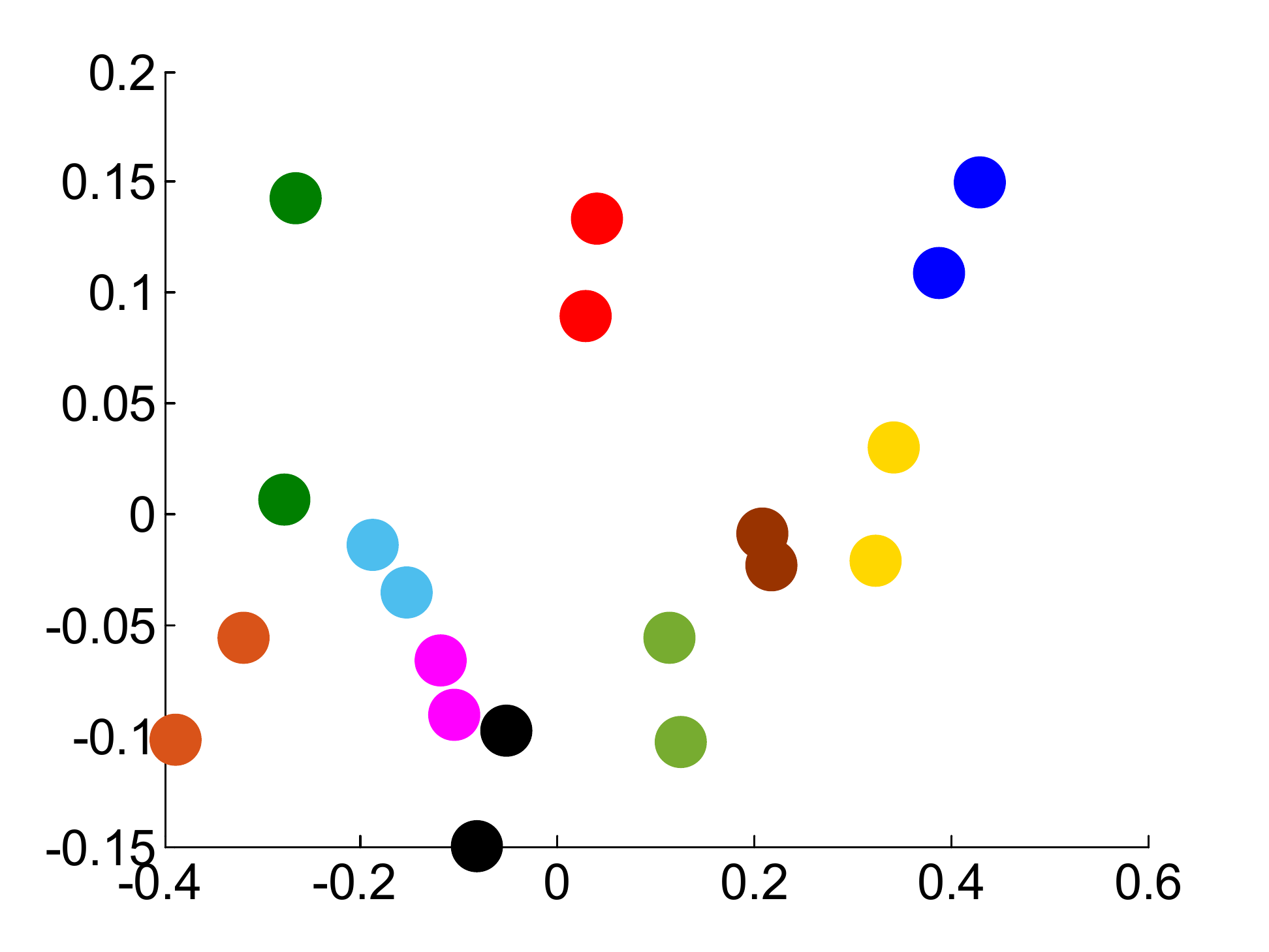}
}
} 
\caption{Illustration of the effectiveness of TDL, where 10 different persons in PRID 2011 dataset were selected for demonstration. Points of different colors indicate different persons. The left are the person data points in the original 2-D space and the right is the projected person data points in the 2-D space learned by TDL. The projection matrix $\mathbf{L}$ is obtained by decomposing matrix $\mathbf{M}$ into $\mathbf{M}=\mathbf{L}^{\top}\mathbf{L}$.}
\label{fig:PRIDtrain}
\end{figure}

To visualize the effectiveness of TDL, a comparison between the data distributions of the original feature space and the latent feature space learned by TDL is shown in Figure~\ref{fig:PRIDtrain}. The change of distribution indicates that the input data samples of the same person are ambiguous, while TDL does reduce the ambiguities and the data distribution is much favorable for classification.




\subsection{Optimization}

To simplify our notation, we denote the outer product of pairwise differences by
\begin{equation}\label{eq:outer}
\begin{aligned}
\mathbf{X}_{i,j}=(\vec{x}_i-\vec{x}_j)(\vec{x}_i-\vec{x}_j)^{\top}.
\end{aligned}
\end{equation}
Based on Eq. (\ref{eq:outer}), we can reformaulate $\mathcal{D}(\vec{x}_i,\vec{x}_j)$ as follows:
\begin{equation}\begin{aligned}
\mathcal{D}(\vec{x}_i,\vec{x}_j) = \mathit{tr(\mathbf{M}\mathbf{X}_{i,j})}，
\end{aligned}\end{equation}
and therefore we can reformulate the objective function Eq.(\ref{eq:toppush_criterion}) as:
\begin{equation}\label{eq:criterion_new}
\begin{split}
f(\mathbf{M}){}=(1- & \alpha)\sum_{\vec{x}_i,\vec{x}_j,y_i=y_j}\mathit{tr(\mathbf{M}\mathbf{X}_{i,j})} {} \\
+\alpha\sum_{\vec{x}_i,\vec{x}_j,y_i=y_j}
& \max\bigl\{\mathit{tr(\mathbf{M}\mathbf{X}_{i,j})}-\min_{y_k\not=y_i}\mathit{tr(\mathbf{M}\mathbf{X}_{i,k})} + \rho,0 \bigr\}.
\end{split}
\end{equation}

Our model applies a stochastic gradient descent projection method to compute an optimized positive semi-definite matrix $\mathbf{M}$ in Eq.(\ref{eq:criterion_new}).
In particular, at the $t$-th iteration,
Eq.(\ref{eq:criterion_new}) is piecewise linear with respect to $\mathbf{M}$. At step $t$, given $\mathbf{M}=\mathbf{M}_t$, we define a set of indices 
$(i,j,k)\in\mathcal{N}(\mathbf{M}_t)$,
if and only if the indices $(i,j,k)$ trigger the second term of Eq.(\ref{eq:criterion_new}). The stochastic gradient $\mathbf{G}_t$ of $f(\mathbf{M})$ at step $t$ is computed by:
\begin{equation}\small
\begin{aligned}
\mathbf{G}_t=\frac{\partial f}{\mathbf{M}}|_{\mathbf{M}=\mathbf{M}_t}=&(1- \alpha)\sum_{i,j} \mathbf{X}_{i,j}\\
&+\alpha \sum_{(i,j,k)\in \mathcal{N}(\mathbf{M}_t)}(\mathbf{X}_{i,j}-\mathbf{X}_{i,k}).
\end{aligned}\end{equation}

The optimization of Eq.(\ref{eq:criterion_new}) must satisfy the constraint that the matrix $\mathbf{M}_{t+1}$ remains positive semi-definite.
For this purpose, we project $\mathbf{M}_{t+1}$ onto the cone of all positive semi-definite matrices
$\mathcal{P}_+$ after each gradient descent step.
To be specific, we first perform the eigen-decomposition on $\mathbf{M}_{t+1}$:
\begin{equation}
\begin{aligned}\label{eq:semi-positive}
\mathbf{M}_{t+1}=\mathbf{V}_{t+1}\mathbf{D}_{t+1}\mathbf{V}_{t+1}^\top.
\end{aligned}
\end{equation}
In order to apply the projection, we will update the diagonal matrix $\mathbf{D}_{t+1}$ by removing all the negative eigenvalues, and then reconstruct $\mathbf{M}_{t+1}$ by Eq.(\ref{eq:semi-positive}).

The algorithm is summarized in \textbf{Algorithm 1}. We denote the gradient step size by $\lambda>0$. In practice, it worked starting with $\lambda=1e-03$.
Then, at each iteration, we increased $\lambda$ by a factor of 1.01 if the loss function decreased and decreased $\lambda$ by a factor of 0.5 if the loss function increased.
\renewcommand{\algorithmicrequire}{\textbf{Initialize:}}
\renewcommand{\algorithmicensure}{\textbf{Output:}}
\begin{algorithm}[t]         
\caption{ The Optimisation Algorithm for TDL.}
\begin{algorithmic}[1]
\REQUIRE  ~~\\                          
    Initialize metric with the identity matrix $\mathbf{M}_0:=\mathbf{I}$;\\
    The triggered set $\mathcal{N}(\mathbf{M}_0):=\{\}$ ;\\
    The gradient $\mathbf{G}_t:=(1- \alpha)\sum_{i,j} \mathbf{X}_{i,j}$ ;\\
    The counter $t:=0$.
\WHILE {(not converged)}
\STATE Search the smallest between-class distance by Eq.(3).
\STATE Construct the triggered set $\mathcal{N}(\mathbf{M}_{t})$ by indices $(i,j,k)$ determined by the second term of Eq.(4).
\STATE Compute $\mathbf{G}_t$ by Eq.(9).
\STATE Compute $\mathbf{M}_{t+1}:= \mathbf{M}_{t}-\lambda\mathbf{G}_t$.
\STATE Project $\mathbf{M}_{t+1}$ onto the cone of all positive semi-definite matrices $\mathcal{P}_+(\mathbf{M}_{t+1})$.
\STATE $t:=t+1$.
\ENDWHILE
\RETURN ~~                          
 $\mathbf{M}_t$.
\end{algorithmic}
\label{code:algorithm}
\end{algorithm}

\noindent\textbf{Matching.}
The learned metric can be exploited to perform person re-id by matching a probe person video sequence $\vec{x}_p$ against a gallery set $\{\vec{x}_g\}$ in another camera view. The distance between a probe video sequence $\vec{x}_p$ and a gallery video sequence $\vec{x}_g$ is computed by
\begin{equation}\begin{aligned}
\mathcal{D}(\vec{x}_p,\vec{x}_g)=(\vec{x}_p-\vec{x}_g)^\top\mathbf{M}(\vec{x}_p-\vec{x}_g).
\end{aligned}\end{equation}

\section{Experiments}

\subsection{Datasets and settings}

\begin{figure*}[t]
\centering{
\subfigure[PRID 2011]
{
    \label{fig:dataset_PRID}
   \includegraphics[width=0.48\linewidth,height=0.12\linewidth]{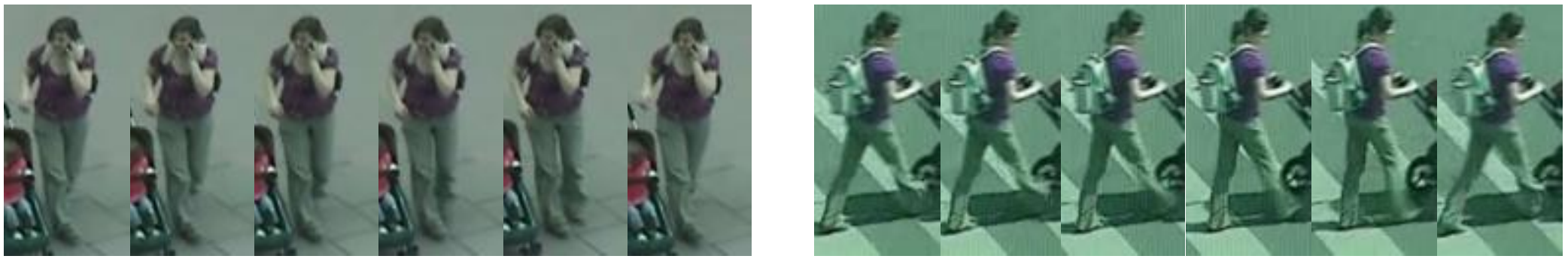}
}
\subfigure[iLIDS-VID]
{
    \label{fig:dataset_iLIDS}
   \includegraphics[width=0.48\linewidth,height=0.12\linewidth]{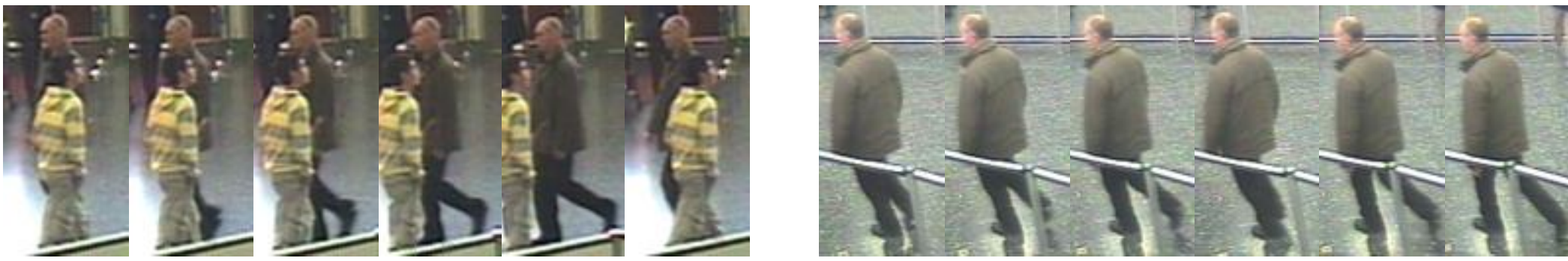}
}
} 
\caption{Example pairs of image sequences of the same person appearing in different camera views.}
\end{figure*}

\noindent\textbf{Datasets.}
Our experiments were conducted on two publicly available video datasets for video-based person re-id: the PRID 2011 dataset \cite{hirzer2011person} and the iLIDS-VID dataset \cite{wang2014person}.
The PRID 2011 dataset consists of video pairs recorded from two different but static surveillance cameras.
385 persons were recorded in camera view A, and 749 persons in camera view B. Among all persons, 200 persons were recorded in both camera views.
Each video is comprised of 5 to 675 image frames, with an average of 100 for each.
To guarantee the effective length of the video, we selected 178 persons with more than 27 frames in our experiments.
This dataset was captured in uncrowded outdoor scenes with relatively simple and clean background and rare occlusions, and several different poses of person are available in each camera view (Figure~\ref{fig:dataset_PRID}~).
The iLIDS-VID dataset contains 600 video of 300 randomly sampled people. Each person has one pair of video from two camera views.
Each video is comprised of 23 to 192 image frames, with an average of 73 for each.
Compared with the PRID 2011 dataset, it was captured in an airport arrival hall under a multi-camera CCTV network.
The challenges of this dataset largely lie in clothing similarities, lighting and viewpoint changes across camera views, complicated background and occlusions (Figure~\ref{fig:dataset_iLIDS}~).

\begin{table*}
\begin{center}
\footnotesize
\newcommand{\tabincell}[2]{\begin{tabular}{@{}#1@{}}#2\end{tabular}}
\renewcommand\arraystretch{1.1}
  \centering
 \setlength{\tabcolsep}{6pt}
\begin{tabular}{|c|c|c|c|c|c|c|c|c|}
\hline
 \multirow{2}{*}{Methods}   &              \multicolumn{ 4}{c|}{PRID 2011}    &                    \multicolumn{ 4}{c|}{iLIDS-VID} \\\cline{2-9}
    & Rank-1      & Rank-5      & Rank-10     & Rank-20     & Rank-1      & Rank-5      & Rank-10     & Rank-20 \\
\hline
TDL &\textbf{56.74}  &\textbf{80.00}  &\textbf{87.64}  &\textbf{93.59}  &\textbf{56.33} &\textbf{87.60} &\textbf{95.60} &\textbf{98.27}  \\
\hline
SDALF \cite{farenzena2010person}
&5.2          &20.7         &32.0         &47.9         &6.3          &18.8         &27.1         &37.3     \\
Salience \cite{zhao2013unsupervised}
&25.8         &43.6         &52.6         &62.0         &10.2         &24.8         &35.5         &52.9     \\
RPRF \cite{li2015multi}
&19.3         &38.4         &51.6         &68.1         &14.5         &29.8         &40.7         &58.1     \\
SRID \cite{karanam2015sparse}
&35.1         &59.4         &69.8         &79.7         &24.9         &44.5         &55.6         &66.2     \\
DVDL \cite{karanam2015iccv}
& 40.6        & 69.7        & 77.8        & 85.6        & 25.9        & 48.2        & 57.3        & 68.9      \\
Color\&LBP+DVR \cite{wang2014person}
& 37.6        & 63.9        & 75.3        & 88.3        & 34.5        & 56.7        & 67.5        & 77.5
\\
\hline
\end{tabular}
\end{center}
\caption{Comparison with the state-of-the-art methods on PRID 2011 and iLIDS-VID datasets. Results are shown as matching rates ($\%$) at Rank = 1, 5, 10, 20. Best results are in boldface font.}
\label{table:vs.the state of the art}
\end{table*}

\noindent\textbf{Settings.}
In our experiments, we adopted a single-shot experiment setting. All datasets were randomly divided into training set and testing set by half so that there were $p=89$ and $p=150$ individuals in the testing sets of PRID 2011 and iLIDS-VID respectively. In the testing stage, the videos from one camera were used as the gallery set while the ones from  another camera as the probe set. The cumulative matching characteristic (CMC) curve is used to measure the performance of each method on each dataset. A rank $k$ matching rate indicates the accuracy of the matching between the probe video $\vec{x}_p$ and the gallery videos $\{\vec{x}_g\}_{g=1}^k$ in the top $k$ rank list. To obtain statistically reliable results, we repeated the procedure 10 times and reported the average results.

\subsection{Feature Extraction}\label{feature}

To obtain more abundant and robust features for representing a person video, we explored a combined person video feature representation. We expressed each sample with appearance feature on image frame level and space-time feature on video level.
Specifically, at the image frame level, each frame of the person video was resized to $128\times48$ pixels and divided into patches with size $8\times 16$ with $50\%$ overlap both in the horizontal and vertical directions.
That is to say, there were $155$ patches for extracting color histograms and LBP features \cite{hirzer2012relaxed}.
For each patch, histograms of color channels in HSV and LAB color spaces and LBP descriptor were computed.
All the appearance feature descriptors within the image frame were concatenated together to form a 1705-dimensional feature vector.
At the video level, we extracted a 1200-dimensional HOG3D feature vector for each person video \cite{klaser2008spatio}.
In the end, we described the whole person video using a 2905-dimensional vector by connecting this HOG3D feature with average pooling of color histograms and LBP features over all image frames of the video.

\begin{figure*}[t]
\centering{
\subfigure[PRID 2011]
{
   \label{fig:rate-PRID}
   \includegraphics[width=0.48\linewidth,height=0.30\linewidth]{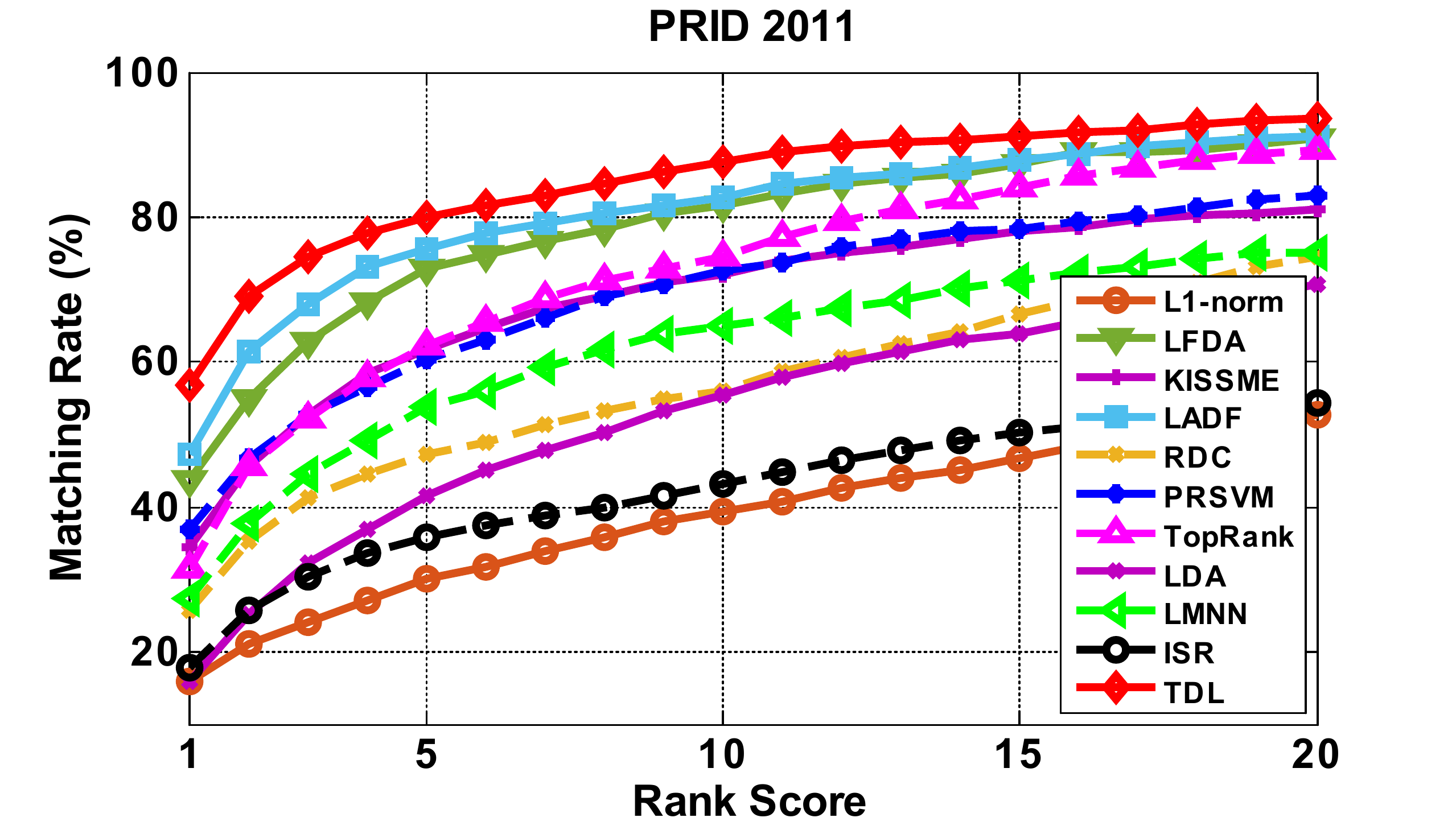}
}
\subfigure[iLIDS-VID]
{
   \label{fig:rate-iLIDS}
   \includegraphics[width=0.48\linewidth,height=0.30\linewidth]{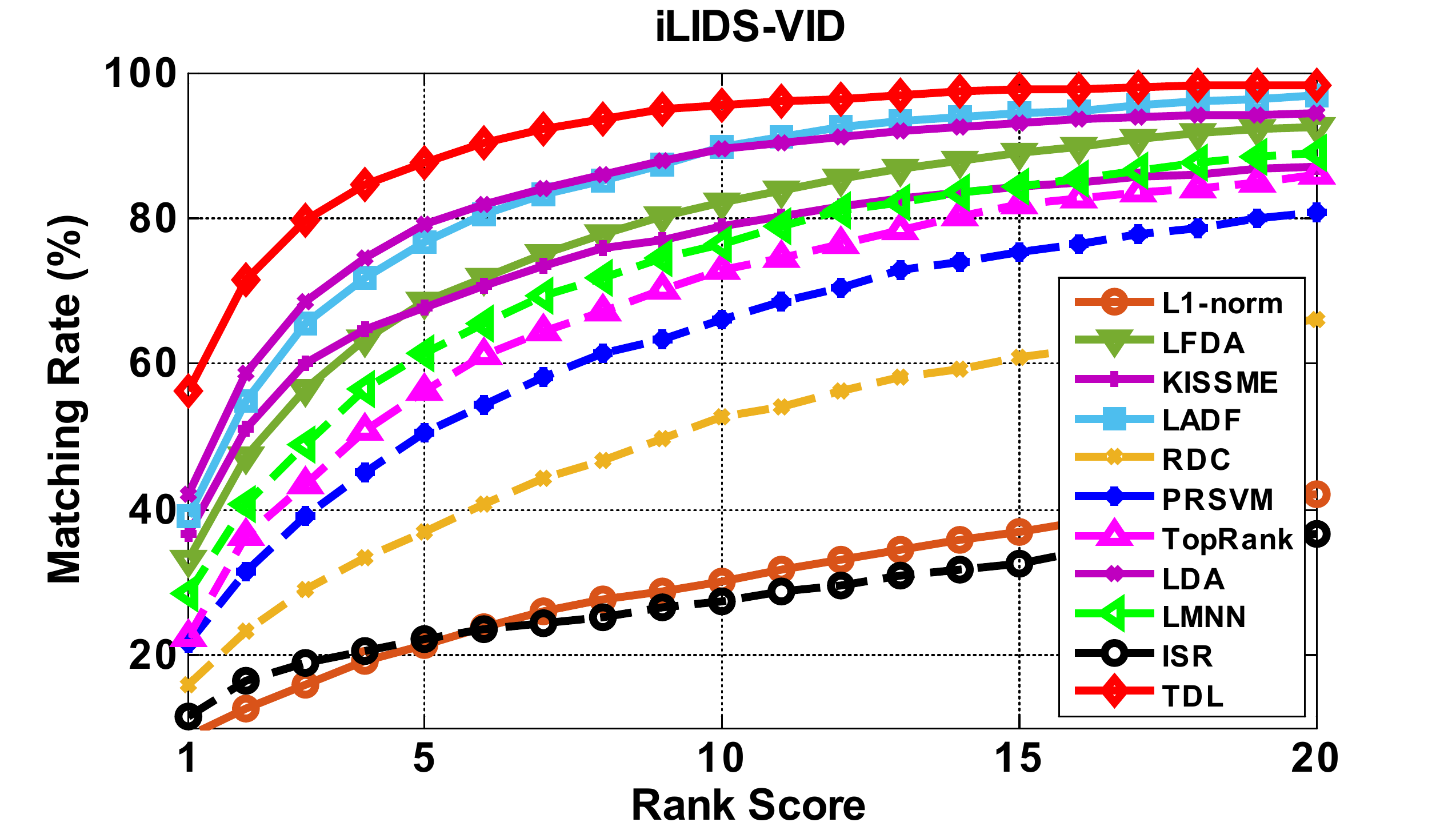}
}
}  
\caption{Video-based matching rates ($\%$) of different methods on PRID 2011 and iLIDS-VID. }
\label{fig:vs.related methods}
\end{figure*}

\subsection{Evaluation of Comparison}

\subsubsection{Comparison with the State-of-the-art Methods}

In Table \ref{table:vs.the state of the art}, we reported the comparison of our proposed TDL model with the existing six state-of-the-art video-based person re-id methods on PRID 2011 and iLIDS-VID datasets, including SDALF \cite{farenzena2010person}, Salience \cite{zhao2013unsupervised}, RPRF \cite{li2015multi}, SRID \cite{karanam2015sparse}, DVDL \cite{karanam2015iccv} and Color\&LBP+DVR \cite{wang2014person}. 
\modify{DVDL is a dictionary learning method based on multi-shot re-id datasets.
DVR is a method based on ranking model, which also selects discriminative video fragment from a candidates pool in the training process.
The results show clearly that with the proposed TDL model, the matching performance on both datasets is improved significantly.
For instance, on iLIDS-VID dataset, our TDL improved the Rank-1 matching rate by $21.8\%$ compared to the second best method Color$\&$LBP+DVR.} 

\modify{Another interesting but indeed fact can be observed is that TDL outperformed others much better on iLIDS-VID. We examined that this is probably because more intra-class distances could be much larger than inter-class ones under more occlusions on iLIDS-VID. While the compared distance models do not explicitly and directly quantify the relation between each intra-class distance  and the related minimum inter-class distance, the proposed TDL employs the top-push strategy and makes the distance model quantify more effective features and thus performs more stably.}

\subsubsection{Comparison with Related Methods}

\begin{table*}
\footnotesize
\newcommand{\tabincell}[2]{\begin{tabular}{@{}#1@{}}#2\end{tabular}}
\renewcommand\arraystretch{1.1}
  \centering
 \setlength{\tabcolsep}{8pt}
\begin{tabular}{|l|c|c|c|c|c|c|c|c|c|}
\hline
\multirow{2}{*}{Settings} & \multirow{2}{*}{Methods}    &\multicolumn{4}{c|}{PRID 2011}    &\multicolumn{4}{c|}{iLIDS-VID} \\\cline{3-10}
 &Rank   & Rank-1      & Rank-5      & Rank-10     & Rank-20     & Rank-1      & Rank-5      & Rank-10     & Rank-20 \\\hline
~ &TDL &\textbf{56.74} &\textbf{80.00} &\textbf{87.64} &\textbf{93.59}  &\textbf{56.33} &\textbf{87.60} &\textbf{95.60} &\textbf{98.27} \\\cline{2-10}
~ &L1-norm &15.84	&30.00	&39.33	&52.70  &8.90   &21.40  &30.07  &42.07  \\
~ &LFDA  \cite{pedagadi2013local} &43.70  &72.80  &81.69  &90.89  &32.93  &68.47  &82.20  &92.60  \\
~ &KISSME   \cite{koestinger2012large} &34.38	&61.68	&72.13	&81.01  &36.53	&67.80	&78.80	&87.07  \\
\textbf{Video-based} &LADF   \cite{li2013learning} &47.30	&75.50	&82.69	&91.12  &39.00	&76.80	&89.00	&96.80  \\
\textbf{matching} &RDC   \cite{zheng2013reidentification} &25.62	&47.30	&56.07	&74.38  &15.80	&36.93	&52.60	&66.00  \\
~ &PRSVM  \cite{prosser2010person} &36.97	&60.45	&72.47	&83.03  &21.53	&50.60	&66.00	&80.80  \\
~ &ISR  \cite{lisanti2014person} & 17.64  &35.84   &43.03   &54.38  &11.60   &22.13   &27.40   &36.67 \\
~ &TopRank \cite{li2014top} &31.69	&62.24	&75.28	&89.44  &22.53	&56.13	&72.73	&85.93  \\
~ &LDA \cite{fukunaga2013introduction} &15.84  &41.46	&55.51	&70.67  &42.06  &79.13  &89.40  &94.47  \\
~ &LMNN   \cite{weinberger2009distance} &27.19	&53.71	&64.94	&75.17  &28.33	&61.40	&76.47	&88.93  \\\hline
\hline
~ &TDL   &\textbf{30.22}  & 59.10  &\textbf{74.04}  &\textbf{88.43}  & 9.81  & 27.52  &\textbf{46.10}  &\textbf{62.19} \\\cline{2-10}
~ &L1-norm &12.36	&29.44	&40.56	&56.40		&3.67	&10.33	&16.03	&26.93  \\
~ &LFDA    \cite{pedagadi2013local} &26.40	&56.07	&69.89	&81.12		&7.80	&23.93	&36.47	&50.80  \\
\textbf{Multiple image} &KISSME   \cite{koestinger2012large} &28.54	&\textbf{59.78}	&72.13	&83.26		&\textbf{10.67}	&\textbf{28.33}	&39.80	&57.00   \\
\textbf{frames matching} &LADF   \cite{li2013learning} &8.20	&20.45	&29.89	&42.25		&4.33	&14.00	&21.20	&32.13  \\
~ &ISR  \cite{lisanti2014person} &10.50	&20.83	&31.83	&44.17  &8.04	&20.50	&31.33 &43.50 \\
~ &LDA \cite{fukunaga2013introduction} &27.64	&58.09	&69.66	&82.47		&10.27	&27.40	&39.80	&55.27  \\
~ &LMNN   \cite{weinberger2009distance} &14.38	&38.09	&50.22	&67.19		&4.47	&13.20	&21.60	&35.47  \\\hline

\end{tabular}
\vspace{0.2cm}
\caption{Comparison with related methods on PRID 2011 and iLIDS-VID datasets. The matching rate ($\%$) at Rank $i$ means the accuracy of the matching within the top $i$ gallery classes.}
\label{table:vs.related methods}
\vspace{-0.3cm}
\end{table*}

There are several existing distance/subspace learning models usually applied for person re-id. For fair comparison, all compared methods used the same feature representation of person videos described in Sec.~\ref{feature}.
We first compared our TDL with representative rank/distance/subspace learning methods for video-based matching, \eg TopRank \cite{li2014top}, linear discriminant analysis (LDA) \cite{fukunaga2013introduction} and LMNN \cite{weinberger2009distance}.
Our results (Figure~\ref{fig:vs.related methods} and Table~\ref{table:vs.related methods}) show clearly that the proposed TDL model obtains better matching rates than these methods. More specifically, on PRID 2011 dataset, the Rank-1 matching rate is 56.74\% for TDL, whilst 31.69\% for TopRank, 15.84\% for LDA, and 27.19\% for LMNN.
These results show that these related methods performed poorly for video-based person re-id.
As seen from the comparison video-based matching results in Figure~\ref{fig:vs.related methods} and Table~\ref{table:vs.related methods}, the improvement was particularly significant on iLIDS-VID dataset, which is more challenging due to more ambiguities caused by occlusions and illumination. With the top-push constraint, the ambiguities can be better removed.

The video-based matching results of several representative still-image-based person re-id methods are also shown in Figure~\ref{fig:vs.related methods} and Table~\ref{table:vs.related methods}, including L1-norm, LFDA \cite{pedagadi2013local}, KISSME \cite{koestinger2012large}, LADF \cite{li2013learning}, PRSVM \cite{prosser2010person}, RDC \cite{zheng2013reidentification} and ISR \cite{lisanti2014person}.
One can observe that our TDL model always outperformed all the compared re-id methods on both datasets. The improvement is particularly significant on iLIDS-VID dataset, and TDL is 17.33\% higher than the best compared the method at Rank-1.
In addition, among these six re-id methods, RDC is closely related to our model, but RDC is limited by the scale of relative comparison. In our experiments, the computational cost of our model was only 3\% of the one of RDC. These results highlight the effectiveness of the proposed model.

\begin{figure*}[t]
\centering{
\subfigure[PRID 2011]
{
   \includegraphics[width=0.46\linewidth,height=0.28\linewidth]{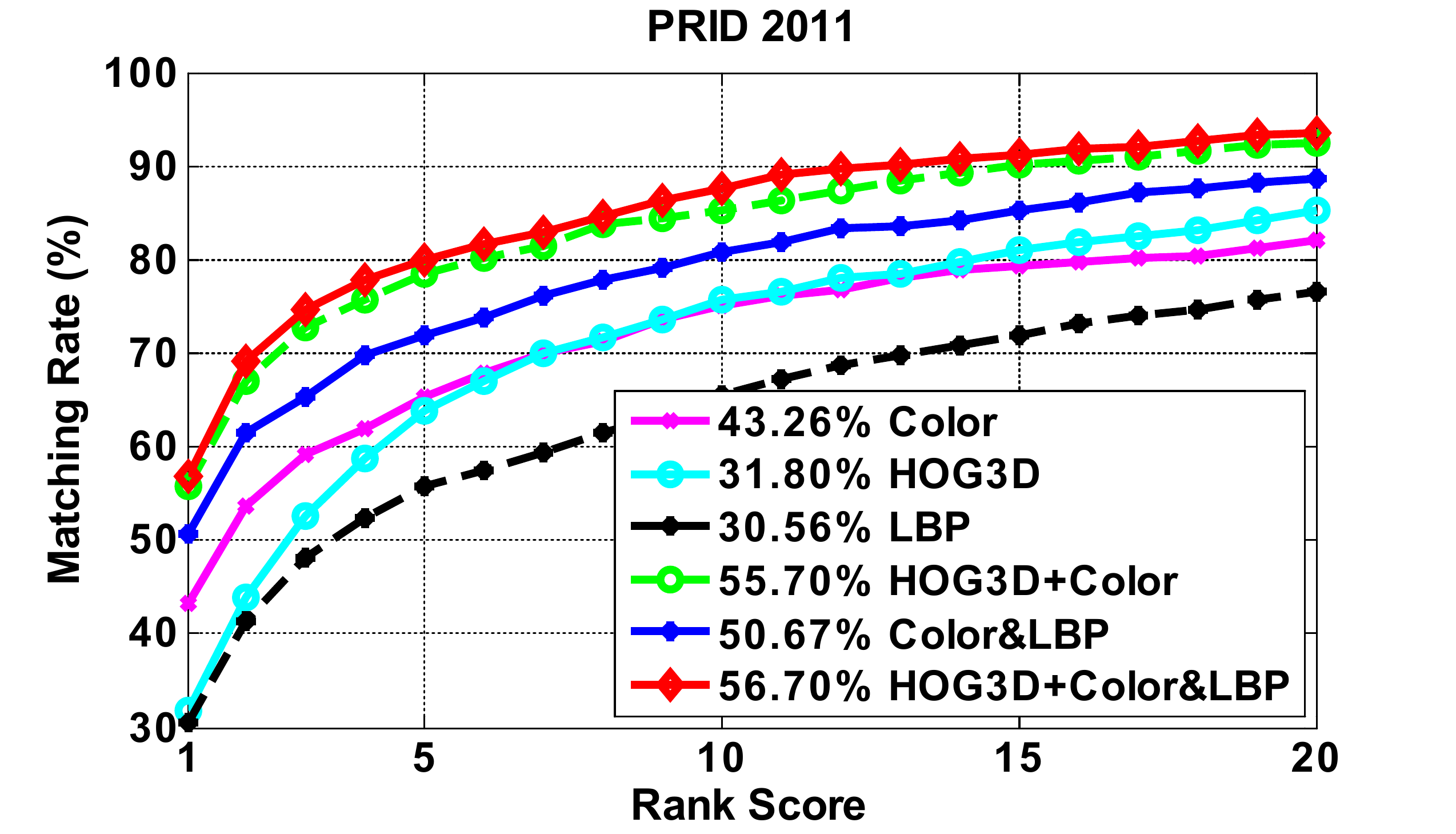}
}
\subfigure[iLIDS-VID]
{
   \includegraphics[width=0.46\linewidth,height=0.28\linewidth]{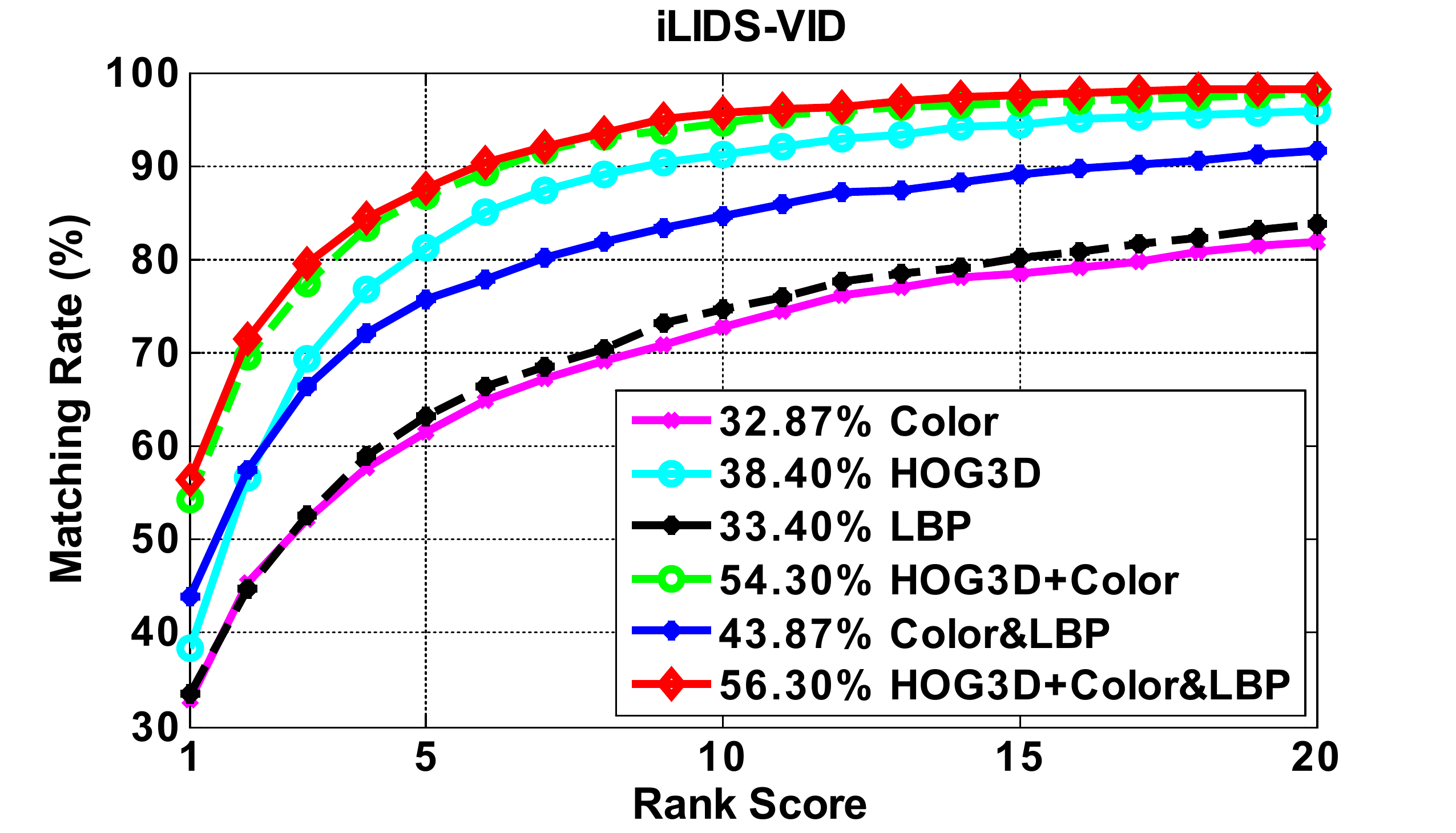}
}
}  
\caption{Evaluation of different feature components in TDL on PRID 2011 and iLIDS-VID. The rank 1 matching rate of each method is provided in the legend.}
\label{fig:components}
\vspace{-0.3cm}
\end{figure*}

One may wonder when using multiple image frames, whether existing still-image-based methods can achieve better performance. To answer the question, in this section, we adopted a multi-frame setting to conduct the experiments, in which 5 images of each person were randomly selected from all frames as gallery.
We used the combined appearance features (Color\&LBP\&HOG) \cite{li2015iccv} as representation of still image frames.
We performed experiments on the two datasets and the results are also reported in Table~\ref{table:vs.related methods}.

\modify{Since RDC, PRSVM and TopRank suffered from the huge computational cost with increasing size of training set under multi-frame or multi-shot setting, these methods cannot be run on a server with 64GB RAM.
To be more specifical, when conducting iLIDS-VID (consists of 300 persons) under multi-shot setting, not just more persons were involved but also more images were used (10 frames for each person in the training set), so that the number of triplets for relative comparison increases dramatically (more than $10^8$). RDC and PRSVM are designed to utilize all the triplets for training, and it is clear that RDC and PRSVM are costly and not computational trackable.
}

Compared to the video-based matching results, it is evident that all the still-image-based methods performed poorly, worse than their video-feature-based versions. This suggests that space-time video information is an important cue to augment the feature representation for person re-id; that is, video-based matching is more effective than multiple image frames matching.

\subsection{Further Evaluation of TDL}
\subsubsection{Effects of Different Feature Components}
\vspace{-0.1cm}
The feature representation used in our proposed model consists of two components: space-time features (HOG3D) and appearance features (Color\&LBP (pooling)).
In Figure~\ref{fig:components}~, we evaluated the effects of each component respectively. The results show that all of them are effective on their own, and when they are combined, the best performance is achieved. This validates that these feature components are complementary and should be fused.

\vspace{-0.2cm}
\subsubsection{Influence of Parameters}
\begin{figure}[t]
\centering{
{
    \label{fig:r}
   \includegraphics[width=0.47\linewidth,height=0.35\linewidth]{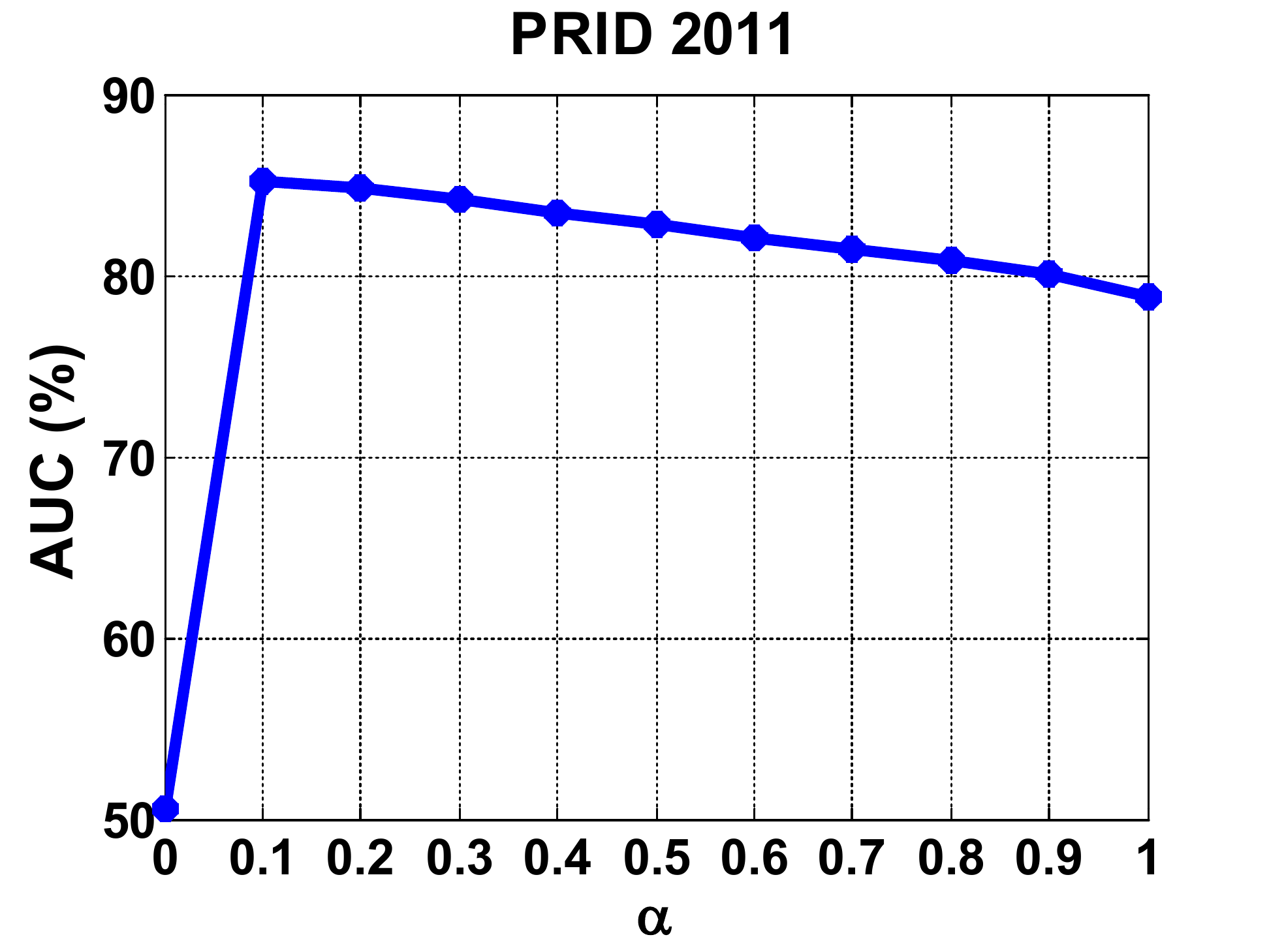}
}
{
    \label{fig:alpha}
   \includegraphics[width=0.47\linewidth,height=0.35\linewidth]{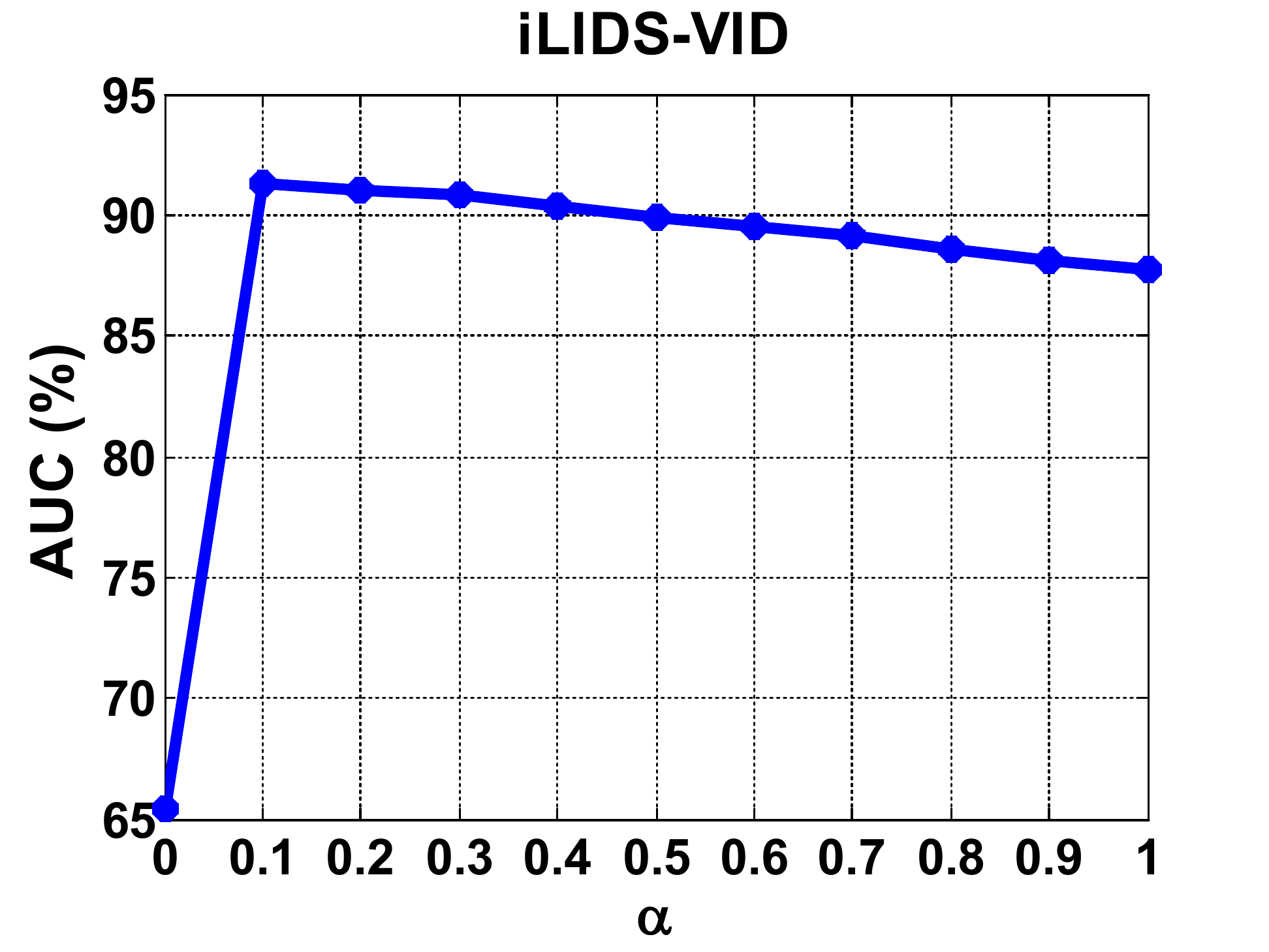}
}
}
\caption{Parameter sensitivity analysis on PRID 2011 and iLIDS-VID}
\label{fig: eval_par}
\vspace{-0.3cm}
\end{figure}

\vspace{-0.05cm}
We implemented our TDL model by selecting the parameter $\alpha$ on PRID 2011 and iLIDS-VID datasets.
The results of area under CMC curve (AUC) were plotted in Figure~\ref{fig: eval_par}~(a) and (b). As illustrated, when $\alpha$ was around 0.1, the model achieved the best result. The figures suggest the performance of using and not using top-push constraint in TDL is distinct. When it is not integrated, the optimization problem Eq. (\ref{eq:criterion_new}) becomes trivial since $\mathbf{M}=\mathbf{O}$ where $\mathbf{O}$ is a zero matrix is the optimal solution which cannot be effective for classification. We also observe that when $\alpha=1$, i.e., discarding the intra-class variation minimization, it will also lead to overfitting in top-push. Thus a proper $\alpha$, for instance 0.1 here is a balance.

\section{Conclusion}
\vspace{-0.1cm}
In this work, we have proposed a top-push distance learning (TDL) model to address the video-based person re-identification problem. While video-based representation contains more abundant space-time information than still-image based representation, there are more ambiguities in video-based features than still-image-based features. So we introduce a top-push constraint to quantify ambiguous video representation. Due to the employment of top-push constraint, the formed distance model can be more effective on top-rank performance of video-based person re-id. This is validated on through extensive experiments conducted on two video datasets including PRID 2011 and iLIDS-VID.

\vspace{-0.4cm}
\section*{Acknowledgments}
\vspace{-0.2cm}
This work was supported in part by the Computational
Science Innovative Research Team Program, Guangdong Provincial Government
of China, in part by the Natural Science Foundation of China under
Grant 61472456, Grant 61522115, and Grant 6151101169,
in part by the Guangzhou Pearl River Science and Technology Rising Star
Project under Grant 2013J2200068, in part by the Guangdong Natural Science
Funds for Distinguished Young Scholar under Grant S2013050014265, and in part by Guangdong Program for Support of Top-notch Young Professionals (No. 2014TQ01X779).

{\small
\bibliographystyle{ieee}
\bibliography{reid}
}

\end{document}